\documentclass{article}

\usepackage[preprint,nonatbib]{neurips_2021}

\usepackage[numbers]{natbib}
\usepackage[utf8]{inputenc} 
\usepackage[T1]{fontenc}    
\usepackage{hyperref}       
\usepackage{url}            
\usepackage{booktabs}       
\usepackage{amsfonts}       
\usepackage{nicefrac}       
\usepackage{microtype}      
\usepackage{xcolor}         

\usepackage{amsmath}  
\usepackage{bm} 
\usepackage{graphicx} 

\usepackage{listings}

\definecolor{codegreen}{rgb}{0,0.6,0}
\definecolor{codegray}{rgb}{0.5,0.5,0.5}
\definecolor{codepurple}{rgb}{0.58,0,0.82}
\definecolor{backcolour}{rgb}{0.95,0.95,0.92}

\lstdefinestyle{mystyle}{
  backgroundcolor=\color{backcolour},   commentstyle=\color{codegreen},
  keywordstyle=\color{magenta},
  numberstyle=\tiny\color{codegray},
  stringstyle=\color{codepurple},
  basicstyle=\ttfamily\footnotesize,
  breakatwhitespace=false,         
  breaklines=true,                 
  captionpos=b,                    
  keepspaces=true,                 
  numbers=left,                    
  numbersep=5pt,                  
  showspaces=false,                
  showstringspaces=false,
  showtabs=false,                  
  tabsize=2
}

\title{Initializing ReLU networks in an expressive subspace of weights}

\author{%
  Dayal Singh \\
  Department of Physics\\
  Indian Institute of Science Education and Research\\
  Pune, India 411008\\
  \texttt{dayal.singh@students.iiserpune.ac.in} \\
   \And
  G J Sreejith \\
  Department of Physics\\
  Indian Institute of Science Education and Research\\
  Pune, India 411008\\
   \texttt{sreejith@acads.iiserpune.ac.in} \\
}

\begin{document}

\maketitle

\begin{abstract}
Using a mean-field theory of signal propagation, we analyze the evolution of correlations between two signals propagating forward through a deep ReLU network with correlated weights. Signals become highly correlated in deep ReLU networks with uncorrelated weights. We show that ReLU networks with anti-correlated weights can avoid this fate and have a chaotic phase where the signal correlations saturate below unity. Consistent with this analysis, we find that networks initialized with anti-correlated weights can train faster (in a teacher-student setting) by taking advantage of the increased expressivity in the chaotic phase. Combining this with a previously proposed strategy of using an asymmetric initialization to reduce dead node probability, we propose an initialization scheme that allows faster training and learning than the best-known initializations.
\end{abstract}

\section{Introduction}

Rectified Linear Unit (ReLU) \cite{Fukushima1969, Fukushima1982} is the most widely used non-linear activation function in Deep Neural Networks (DNNs) \cite{LeCun2015,Ramachandran2018, Nair2010}, applied to various tasks like computer vision \cite{Glorot2011, Krizhevsky2012, He2015}, speech recognition \cite{Maas2013, Lazlo2013, Hinton2012}, intelligent gaming \cite{Silver2016}, and solving scientific problems \cite{Seif2019}. ReLU, $\phi(x) = \max(0, x)$, outperforms most of the other activation functions proposed \cite{Glorot2010}. It has several advantages over other activations. ReLU activation function is computationally simple as it essentially involves only a comparison operation. ReLU suffers less from the vanishing gradients, a major problem in training networks with sigmoid-type activations that saturate at both ends \cite{Glorot2011}. They generalize well even in the overly parameterized regime \cite{Maennel2018}.

Despite its success, ReLU also has a few drawbacks, one of which is the dying ReLU problem \cite{He2015,Trottier2016}. The dying ReLU is a type of vanishing gradient problem in which the network outputs zero for all inputs and is dead. There is no gradient flow in this state. ReLU also suffers from exploding gradient problem, which occurs when backpropagating gradients become large \cite{Boris2018}.

Several methods are proposed to overcome the vanishing/exploding gradient problem; these can be classified into three categories \cite{Lu2020}.
The first approach modifies the architecture, which includes using modified activation functions \cite{Ramachandran2018, He2015, Trottier2016, Clevert2016, Klambauer2017, Hendrycks2016}, adding connections between non-consecutive layers (residual connections)\cite{He2016}, and optimizing network depth and width. The proposed activations are often computationally less efficient and require a fine-tuned parameter \cite{Lu2020}.
The second approach relies on normalization techniques \cite{Ba2016, Ioffe2015, Salimans2016, Ulyanov2016, Wu2018}, the most popular one being batch normalization \cite{Ioffe2015}. Batch normalization prevents the vanishing and exploding gradients by normalizing the output at each layer but with an additional computational cost of up to 30\% \cite{Mishkin2016}. A related strategy involves using the self-normalizing activation (SeLU), which by construction ensures output with zero mean and unit variance \cite{Klambauer2017}.
The third approach focuses on the initialization of the weights and biases. As local (gradient-based) algorithms are used for optimization \cite{Kingma2015, Zeiler2012, Duchi2011}, it is challenging to train deep networks with millions of parameters \cite{Simon2019, Srivastava2015}, and optimal initialization is essential for efficient training \cite{Nesterov2014}. He-initialization \cite{He2015} is a commonly used strategy that uses uncorrelated Gaussian weights with variance $\nicefrac{2}{N}$, where $N$ is the width of the network. Recently, Ref. \cite{Lu2020} proposed Random asymmetric initialization (RAI), which reduces the probability of dead ReLU at the initialization. In this paper, we aim to further improve the initialization scheme for ReLU networks.

A growing body of work has analyzed signal propagation in infinitely wide networks to understand the phase diagram of forward-propagation in DNNs \cite{Saxe2013,Poole2016, Raghu2016, Schoenholz2017, Lee2018, Hayou2018, Saad2018, Li2020,Bahri2020}. We mention a few results for ReLU networks. Ref. \cite{Hayou2018} showed that correlations in input signals propagating through a ReLU network always converge to one. Many other works found that ReLU networks are in general biased towards computing simpler functions \cite{Palma2019, Rahaman2019, Juncai2018, Guillermo2018, Boris2019}, which may account for their better generalization properties even in the overly parameterized regime. However, from their successful application in different domains, one may guess that they should be capable of computing more complex functions. There might be a subspace of the parameters where the network can represent complex functions.

Ref. \cite{Saad2018, Li2020} applied weight and input perturbations to analyze the function space of ReLU networks. They found that ReLU networks with anti-correlated weights compute richer functions than uncorrelated/positively correlated weights. Consistent with this, Ref. \cite{Shang2016} found that ReLU CNN's produce anti-correlated feature matrices after training.
These studies motivated us to analyze the phase diagram of signal propagation in ReLU networks with anti-correlated weights. 

Following the mean-field theory of signal propagation proposed by Ref. \cite{Poole2016}, we found that ReLU networks with anti-correlated weights have a chaotic phase, which implies higher expressivity. In contrast, ReLU networks with uncorrelated weights do not have a chaotic phase. Furthermore, we find that initializing ReLU networks with anti-correlated weights results in faster training. We call it Anti-correlated initialization (ACI). Additional improvement in performance is achieved by incorporating RAI, which reduces the dead node probability. 
This combined scheme, which we call Random asymmetric anti-correlated initialization (RAAI), is the main result of this work and is defined as follows. We pick weights and bias incoming to each node from anti-correlated Gaussian distribution and replace one randomly picked weight/bias with a random variable drawn from a beta distribution. The code to generate weights drawn from the RAAI distribution is given in Appendix \ref{appendix:code}. We analyze the correlation properties of RAAI and show that it performs better than the best-known initialization schemes on tasks of varying complexity. It may be of concern that initialization in an expressive space may lead to overfitting, and we do observe the same for ACI for deeper networks and complex tasks. In contrast, RAAI shows no signs of overfitting and performs consistently better than all other initialization schemes.

We organize the article as follows. First, we contrast the mean-field analysis of ReLU networks with correlated weights with uncorrelated in Section \ref{sec:mft_signal_prop}. Next, Section \ref{sec:RAAI} analyzes the critical properties of correlations in input signals for RAI and RAAI. Then, in Section \ref{sec:tasks}, we describe the various tasks used to validate the performance of different initialization schemes in Section \ref{sec:train_RAAI}. Lastly, Section \ref{sec:discussion} concludes the article.

\section{Mean-field analysis of signal propagation with correlated weights}
\label{sec:mft_signal_prop}

This section presents the mean-field theory of signal propagation (proposed by Ref.~\cite{Poole2016}) in ReLU networks with correlated weights and compares it with uncorrelated weights. Unlike Ref.~\cite{Saad2018, Li2020}, which study perturbation to a ReLU network, we aim to understand the phase diagram of the signal propagation. Furthermore, we provide numerical results to corroborate the mean-field results.

Consider a fully connected neural network with $L$ layers (in addition to the input layer) and $N_l$ nodes in layer $l$. The layer index ranges between $0$ and $L$. For an input signal ${\bm s}^0 = x$, we denote the pre-activation at node $i$ in layer $l$ by $h^l_i(x)$ and activation by $s^l_i(x)$. A signal ($s^{l-1}_1,\; .\;.\; ,s^{l-1}_{N_l}$) at layer $l-1$ propagates to layer $l$ by the rule

\begin{equation*}
    \begin{split}
        & h^l_i(x) = \sum_{j = 1}^{N_{l-1}} w^l_{ij} s^{l-1}_j(x) + b^l_i \;\;\;\;\;\; \text{ where } l \in \{1, L\}\\
        & s^l_i(x) = \phi(h^l_i(x)),
    \end{split}
\end{equation*}

where $\phi$ is the non-linear activation function and $w^l_{ij}, b^l_i$ are the weights and biases. We consider correlations within the set of weights (${\bm w}_i^l$) incoming to each node $i$. The correlated Gaussian distribution is
\begin{equation}
    \label{eqn:corr_weights_dist}
    P({\bm w}^l_1, {\bm w}^l_2, {\bm w}^l_3\dots) = \prod_i^{N_l} \frac{\exp\left(-\frac{1}{2} (\bm{w}^l_i)^T A^{-1} \bm{w}^l_i \right)} {\sqrt{(2\pi)^{N_{l-1}} |A| }},
\end{equation}
with a covariance matrix given by
\begin{equation*}
    \label{eqn:weight_covariance}
    A =  \frac{\sigma^2_w}{N_{l-1}} \left(\mathbb{I} - \frac{k}{1+k}\frac{J}{N_{l-1}} \right) ,
\end{equation*}
where $\mathbb{I}$ is the identity matrix, $J$ is an all-ones matrix, and $k$ parameterizes the correlation strength. Positively correlated and anti-correlated regimes correspond to the regions $-1<k <0$ and $k > 0$, respectively, whereas $k=0$ generates uncorrelated weights. The overall scaling by {\bm $1/N_{l-1}$} in the covariance matrix ensures that the input contribution from the last layer to each node is $\mathcal{O}$(1). The bias is drawn from a Gaussian distribution $b^l_i \sim \mathcal{N}(0, \sigma^2_b)$. Note that weights reaching two different nodes are uncorrelated, and also the bias is uncorrelated with the weights.

To track the layer-wise information flow, consider the squared length and overlap of the pre-activations for two input signals, ${\bm s}^0 = x_1 $ and ${\bm s}^0=x_2$, after propagating to layer $l$ 

\begin{align*}
        & q^l_h(x_a) = \frac{1}{N_l} \sum_{i= 1}^{N_l} \left(h^l_i(x_a)\right)^2 \;\;\;\; \text{ where } a \in \{1, 2\} \text{ and } 1 \leq l \leq L \\
        & q^l_h(x_1, x_2) = \frac{1}{N_l} \sum_{i= 1}^{N_l} h^l_i(x_1) h^l_i(x_2).
\end{align*}


Assuming self averaging, consider an average over the weights and the bias incoming to layer $l$. For simplicity of notations later, we use the same symbol for averaged $q^l_h$. 
\begin{align*}
    & q^l_h(x_a) = \frac{\sigma^2_w}{N_{l-1}} \sum_{j, m = 1}^{N_{l-1}} \left(\delta_{j, m} - \frac{k}{1+k} \frac{1}{N_{l-1}}\right) \phi(h^{l-1}_{j}(x_a)) \phi(h^{l-1}_{m}(x_a)) + \sigma^2_b \\
    & q^l_{h}(x_1, x_2) = \frac{\sigma^2_w}{N_{l-1}} \sum_{j, m = 1}^{N_{l-1}} \left(\delta_{j, m} - \frac{k}{1+k} \frac{1}{N_{l-1}}\right) \phi(h^{l-1}_{j}(x_1)) \phi(h^{l-1}_{m}(x_2)) + \sigma^2_b,
\end{align*}

For large width, each $h^{l-1}_i$ is a weighted sum of a large number of zero-mean random variables. Thus, we expect the joint distribution of $h^{l-1}_i(x_1)$ and $h^{l-1}_i(x_2)$ to converge to a zero-mean Gaussian with a covariance matrix with diagonal entries $q^{l-1}_h(x_1), q^{l-1}_h(x_2)$ and off-diagonal entries $q^{l-1}_h(x_1, x_2)$. On replacing the average over $h^{l-1}_i$ (this is equivalent to considering an average over all previous layers) in the last layer with an average over this Gaussian distribution, we obtain iterative maps for the length and overlap. Specializing to ReLU activation leads to these simplified equations

\begin{equation}
    \label{eqn:maps}
    \begin{split}
     &q^l_h(x) =  \frac{\sigma^2_w }{2} \left(1 - \frac{ k}{1+k} \frac{1}{\pi} \right) q^{l-1}_h(x)  + \sigma^2_b  \\
     &q^l_{h}(x_1, x_2) =  \frac{\sigma^2_w }{2} \left( f(c^{l-1}_h) - \frac{ k}{1+k} \frac{1}{\pi} \right)\sqrt{q^{l-1}_h(x_1)q^{l-1}_h(x_2)} + \sigma^2_b\\
     & f(c^{l-1}_h) =  \frac{c^{l-1}_h}{2} + \frac{c^{l-1}_h}{\pi}  \text{sin}^{-1}(c^{l-1}_h)  + \frac{1}{\pi}\sqrt{1 - (c^{l-1}_h)^2},
    \end{split}
\end{equation}
 where $c^l_{h} = \nicefrac{q^l_h(x_1, x_2)}{\sqrt{q^l_{h}(x_1)q^l_{h}(x_2) }}$ is the correlation coefficient between the two signals reaching layer $l$. We provide the details of the derivation and equations for general activations in Appendix \ref{appendix:ACI}.

Ref. \cite{Poole2016} found that the signal's length reaches its fixed point within a few layers, and the fixed point of the correlation coefficient, $c^*_{h}$, can be estimated with the assumption that $q^l_h(x)$ has reached its fixed point $q^*_h$. We can check that $c^*_h = 1$ is always a fixed point of the recursive map (Equation \ref{eqn:maps}) under this assumption. The stability of the fixed point $c^*_{h} = 1$ is determined by
\begin{equation*}
 \chi_1 \equiv \frac{\partial c^l_{h}}{\partial c^{l-1}_{h}}\Bigr\rvert_{c^{l-1}_h= 1},
 \end{equation*}
which evaluates to $\nicefrac{\sigma^2_w}{2}$.
$\chi_1$ separates the parameter space into two phases \textemdash first, an ordered phase with $\chi_1 < 1$, where the $c^*_{h} = 1$ fixed point is stable; and second, a chaotic phase with $\chi_1 > 1$, where the $c^*_{h}= 1$ fixed point is unstable. $\chi_1 = 1$ defines the phase boundary line.
In the ordered phase, two distinct signals will become perfectly correlated asymptotically. In the chaotic phase, the correlations converge to a stable fixed point below unity. Two closely related signals will eventually lose correlations in this phase. This suggests that initializing the network with parameters ($\sigma^2_w, \sigma^2_b$) at the phase transition boundary (corresponding to an infinite depth of correlations) allows for an optimal information flow through the network \cite{Poole2016,Raghu2016,Schoenholz2017}

\begin{figure}[h]
\centering
\includegraphics[width=0.9\linewidth]{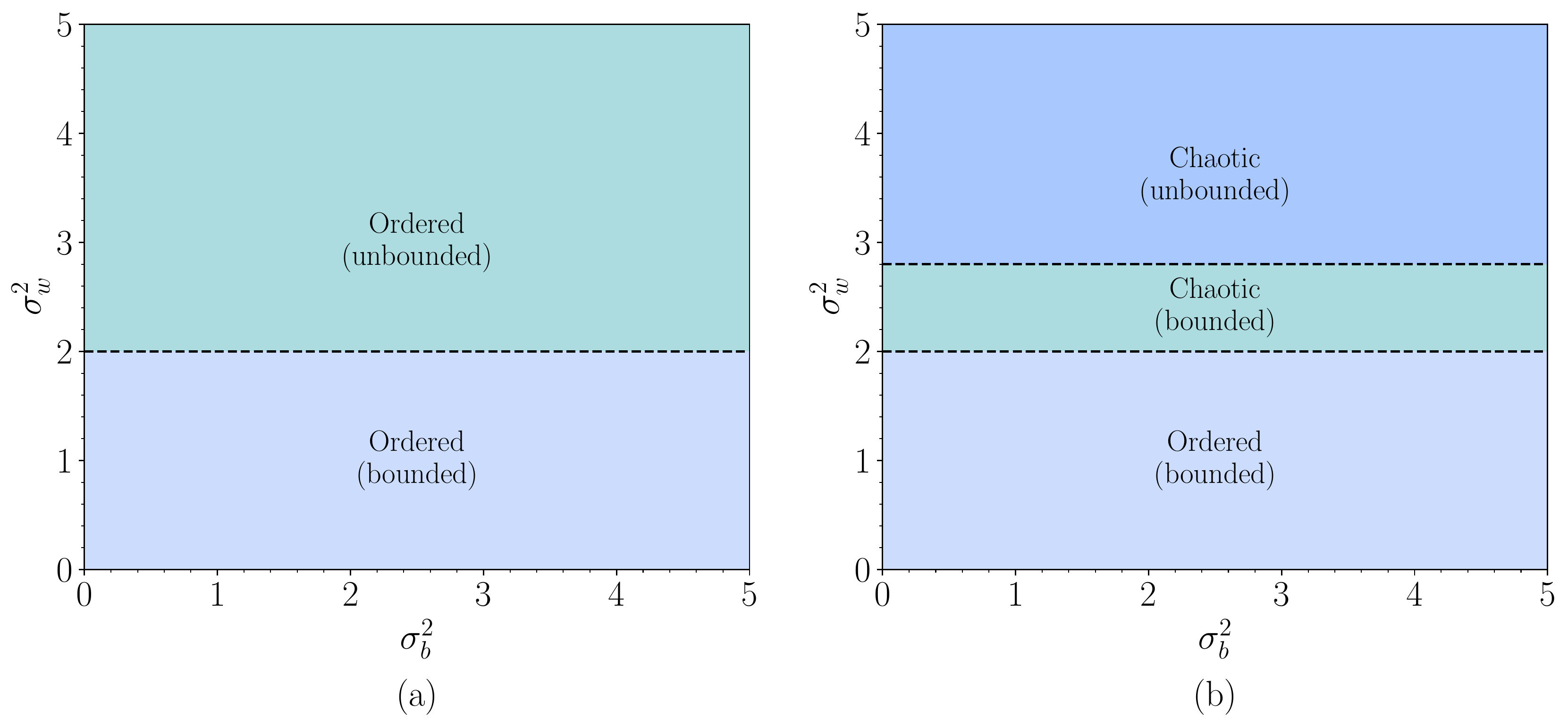}
\caption{Phase diagram for ReLU networks with uncorrelated and anti-correlated Gaussian distributed weights. (a) ReLU networks with uncorrelated weights have two phases. First, a bounded phase where $q^*_h$ is finite and second in which it diverges. The two phases are separated by $\sigma^2_w = 2$. In both phases, any two signals will eventually become correlated. (b) ReLU networks with anti-correlated weights have three phases. In addition to the transition between the bounded and unbounded phases (at $\sigma^2_w = g_k$) there is an order to chaos transition at $\sigma^2_w = 2$. The results are shown for $k = 100$. }
\label{fig:relu_phase_space}
\end{figure}

\begin{figure}[h]
  \centering
  \includegraphics[width=\linewidth]{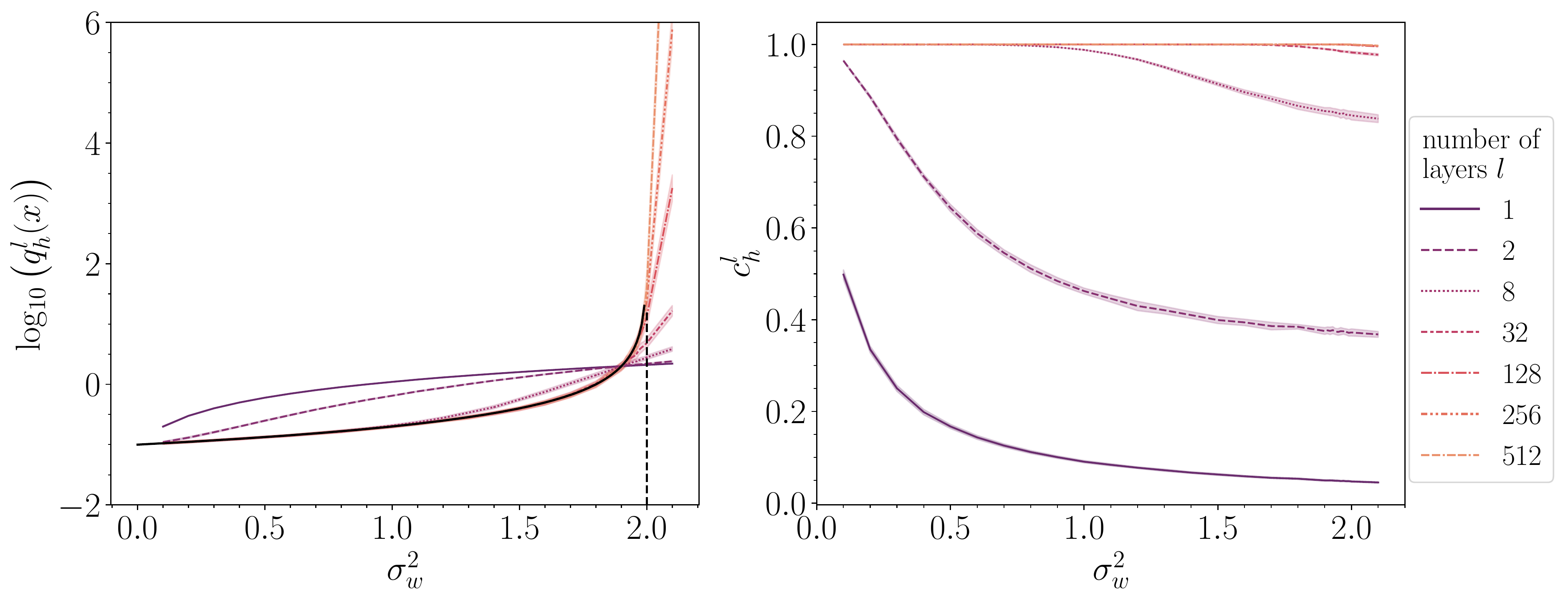}
    \caption{The above plots show the signal's length and correlation coefficient after propagating through $l$ layers in a ReLU network with uncorrelated weights. We estimate the length and correlation coefficient averaged over $M = 1024$ input signals, and 40 networks with a constant width $N = 2048$. The shaded regions denotes the standard deviation. In the first panel, the vertical dashed line indicates the theoretical phase boundary $\sigma^2_w = 2$, and the solid black line denotes the theoretical prediction for the length's fixed point. As the critical boundaries do not depend on the variance of the bias, we show results for $\sigma^2_b = 0.1$ only. We find that $c^l_h \to 1$ for all values of $\sigma^2_w$ and $\sigma^2_b$.
    }
    \label{fig:UI_forward_prop}
\end{figure}

 \paragraph{ReLU networks with uncorrelated weights (k = 0)}~The above analysis is applied assuming $q^*_h$ is finite shows that ReLU networks with uncorrelated weights do not have a chaotic phase, and any two signals propagating through a ReLU network become asymptotically correlated for all values of $(\sigma_w^2, \sigma_b^2)$. In other words, $c^*_h = 1$ is always a stable fixed point.
However, the parameter space can be classified into two phases based on the boundedness of the fixed point $q^*_h$ of the length map (Equation \ref{eqn:maps}) -
first, a bounded phase where $q^*_h$ is finite and non-zero; second, an unbounded phase, where $q^*_h$ is either zero or infinite  \cite{Lee2018,Hayou2019}. The two phases are separated by the boundary $\sigma^2_w = 2$. 
Figure \ref{fig:relu_phase_space}a depicts the phase diagram for ReLU networks with uncorrelated weights.
Note that the analysis of the stability of $c^*_h=1$ fixed point in ReLU networks is valid only in the bounded phase. However, numerical results presented in Figure \ref{fig:UI_forward_prop} indicate that the fixed point remains stable even in the unbounded phase.

\paragraph{ReLU networks with correlated weights}~The phase diagram for ReLU networks with correlated weights can be analyzed similarly. The length is bounded if $\sigma^2_w < g_k = \nicefrac{2}{\left(1 - \frac{k}{1+k} \frac{1}{\pi} \right)}$. Thus, for anti-correlated weights ($k>0$), the boundary $g_k$ moves upwards relative to the $k = 0$ case (see Figure \ref{fig:relu_phase_space}a).
The $c^*_h=1$ fixed point of the correlations is unstable in this additional region of the bounded phase.

\begin{figure}
  \centering
  \includegraphics[width=1\linewidth]{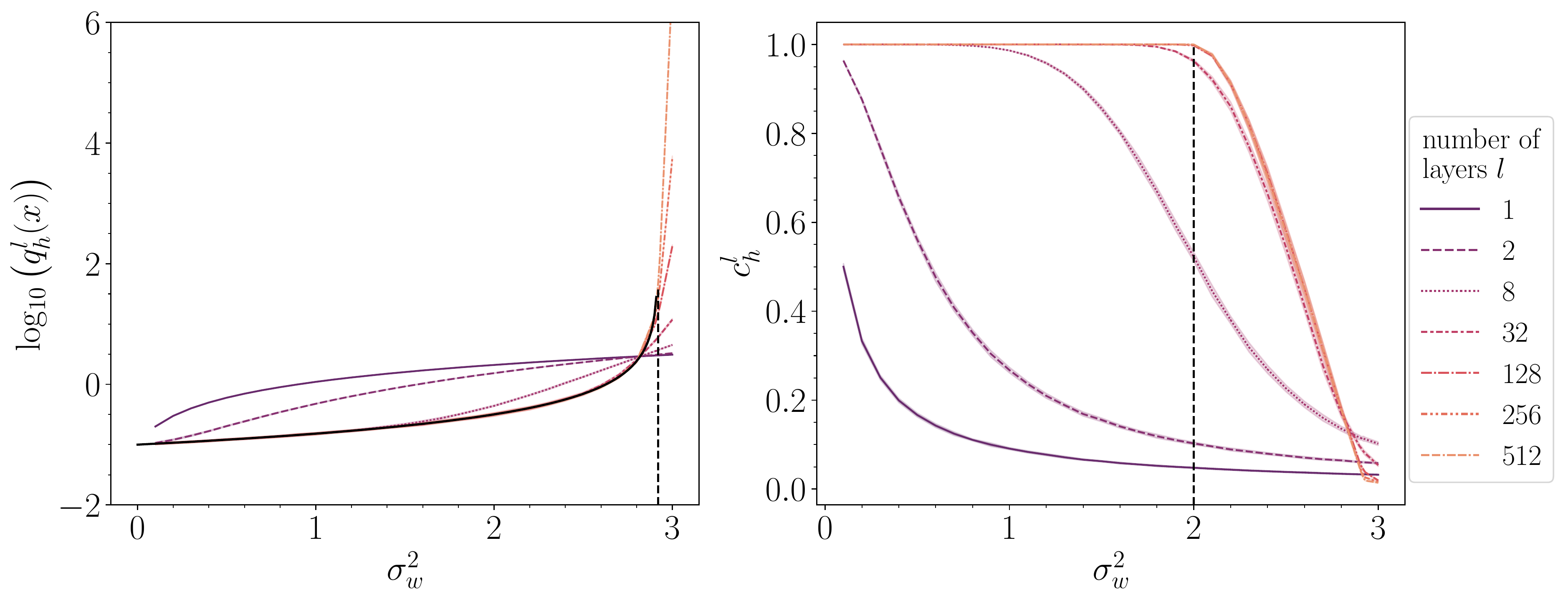}
    \caption{The above plots show the signal's length and correlation coefficient after propagating through $l$ layers in a ReLU network with anti-correlated weights with a correlation strength $k=100$. We estimate the length and correlation coefficient averaged over $M = 1024$ input signals, and 40 networks with a constant width $N = 2048$. The shaded regions denotes the standard deviation. The vertical dashed lines indicate the theoretical phase boundaries at $\sigma^2_w = 2.92$ and $\sigma^2_w = 2.0$ for the length and correlation coefficient. The solid black line in the first panel denotes the theoretical prediction for the length's fixed point. As the critical boundaries do not depend on the variance of bias, we show results only for $\sigma_b^2 = 0.1$. Unlike the case of uncorrelated weights, we find a chaotic region.}
    \label{fig:AI_forward_prop}
\end{figure}

 In summary, anti-correlations induce a bounded chaotic phase in $2 < \sigma^2_w < g_{k}$ (see Figure \ref{fig:relu_phase_space}b).
 We demonstrate these results numerically in Figure \ref{fig:AI_forward_prop} for a correlation strength of $k = 100$. As predicted by the above equations, the stability of the fixed point $c^*_h = 1$ changes at $\sigma^2_w = 2$, and the length diverges at $g_{k=100}=2.92$. In contrast, for positively correlated weights, the length's fixed point boundary shifts downward resulting in a similar phase diagram as uncorrelated weights.

As a result, a ReLU network with anti-correlated weights can be more expressive by taking advantage of a chaotic phase, and it may be beneficial for a ReLU network to remain in this subspace. Thus, we propose initializing ReLU networks with anti-correlated weights at the order to chaos boundary $(\sigma^2_w, \sigma^2_b) = (2, 0)$. We call it Anti-Correlated Initialization (ACI). In Appendix \ref{sec:train_ACI}, we demonstrate that ReLU networks initialized with anti-correlated weights give an advantage over He initializations for a range of tasks.

Many alternatives are proposed to improve ReLU networks~\cite{Trottier2016, Lu2020, Clevert2016}. Of particular interest is Random asymmetric initialization (RAI), which aims to increase expressivity through an independent strategy of reducing the dead node probability. In the next section, we analyze the correlation properties of RAI and then combine it with ACI to propose a new initialization scheme RAAI, which has both a chaotic phase and low dead node probability.

\section{Random Asymmetric Anti-correlated Initialization}
\label{sec:RAAI}
\begin{figure}[h]
  \centering
  \includegraphics[width=1.0\linewidth]{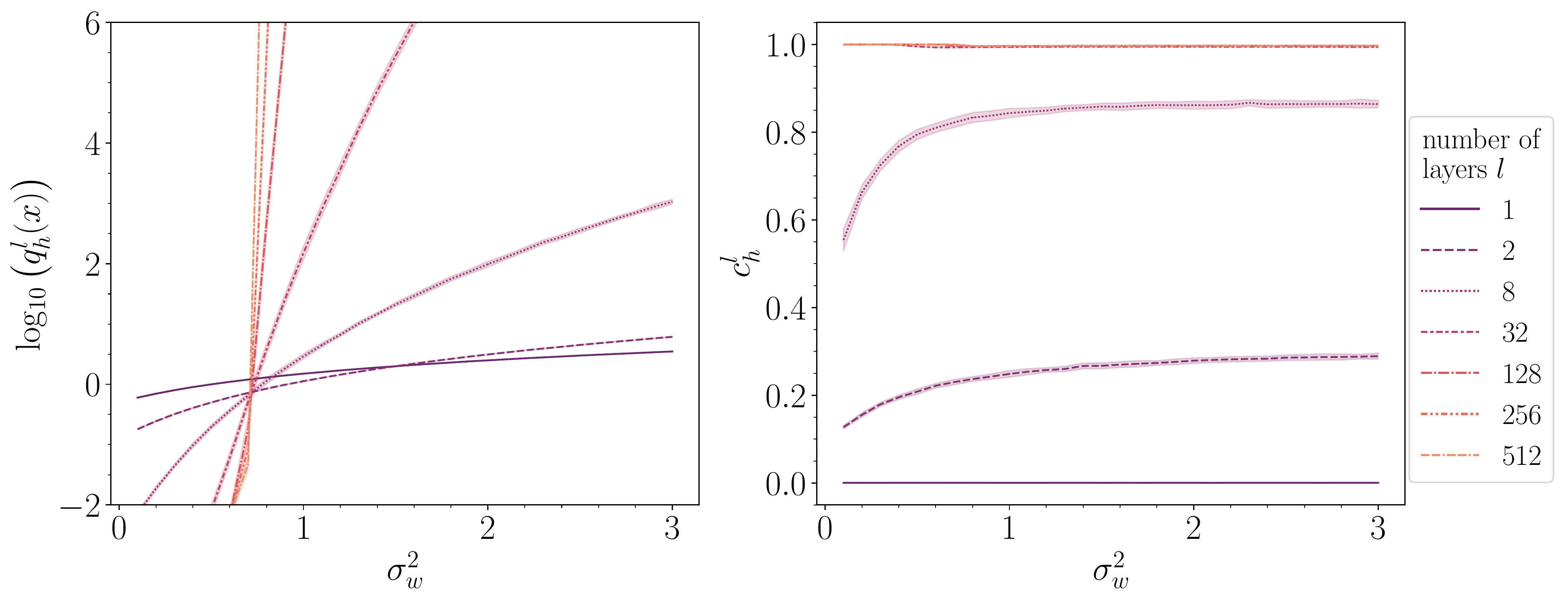}
    \caption{The above plots show the signal's length and correlation coefficient after propagating through $l$ layers in a ReLU network with RAI. We estimate the length and correlation coefficient averaged over $M = 1024$ input signals, and 40 networks with a constant width $N = 2048$. The shaded regions denotes the standard deviation. Similar to Figure \ref{fig:UI_forward_prop}, the chaotic region is absent.}
    \label{fig:RAI_forward_prop}
\end{figure}

We begin by analyzing critical properties of Random asymmetric initialization (RAI) proposed in Ref. \cite{Lu2020} to reduce the dead node probability. For ReLU networks with symmetric distributions for weights and biases, the dead node probability is half. RAI reduces it by initializing one of the weights/the bias incoming to each node from a distribution with positive support (like the beta distribution), resulting in a positive mean for the pre-activations. Ref. \cite{Lu2020} proposes initializing RAI with a variance $\sigma^2_w = 0.36$ to ensure that the signal's length is bounded. We analyzed the correlation properties of RAI and found that $c^*_h = 1$ is always a fixed point of the recursive maps (see Appendix \ref{appendix:RAI_cov_map}).
Deriving the stability condition for the fixed point $c^*_h = 1$ even with the mean-field assumptions is difficult.
However, numerical results presented in Figure \ref{fig:RAI_forward_prop} show that $c^*_h = 1$ is always a stable fixed point (right panel), and the length remains finite for $\sigma^2_w$ up to $0.72$ (left panel). A qualitative picture of the phase diagram can be captured with a few additional assumptions over the mean-field approximation (see Appendix \ref{appendix:stability_corr_map_RAI} for details).

\begin{figure}[h]
  \centering
  \includegraphics[width=1\linewidth]{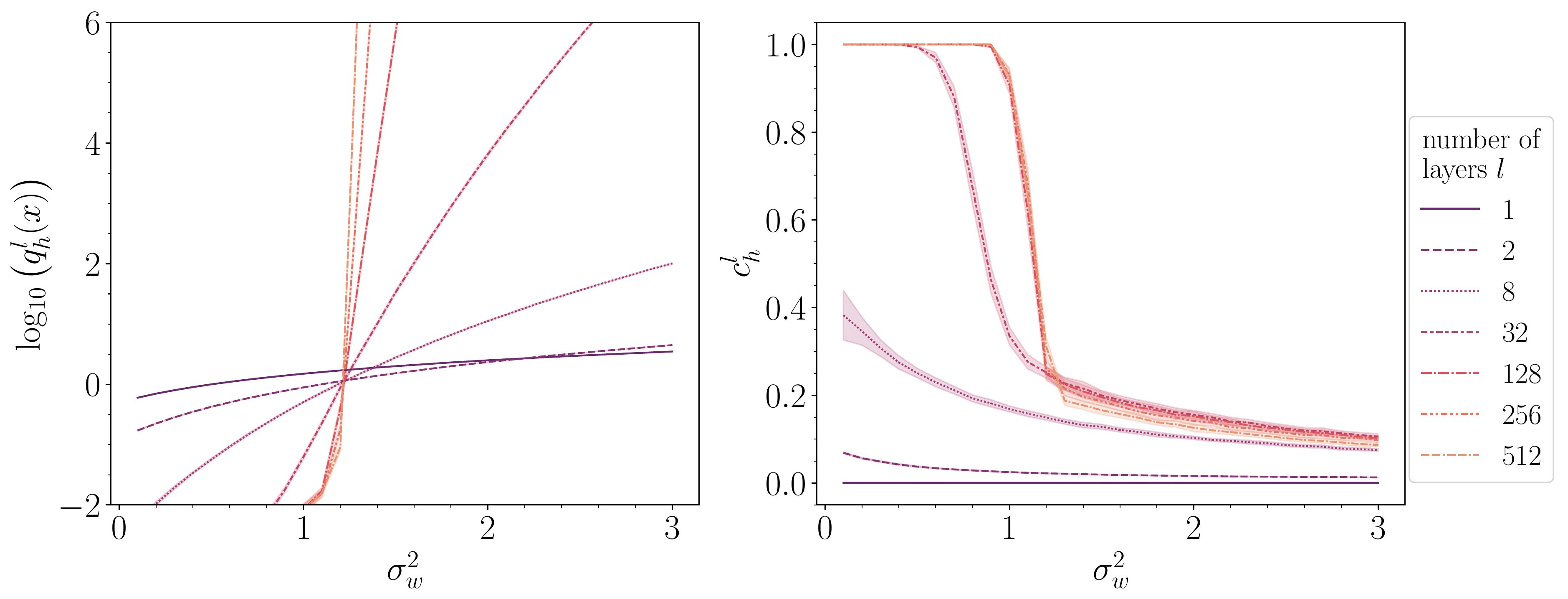}
    \caption{The above plots show the signal's length and correlation coefficient after propagating through $l$ layers in a ReLU network with RAAI. We estimate the length and correlation coefficient averaged over $M = 1024$ input signals, and 40 networks with a constant width $N = 2048$. The shaded regions denotes the standard deviation. Similar to ACI, we find a chaotic region. However, the correlations do not converge to zero even for large $\sigma^2_w$.}
    \label{fig:RAAI_forward_prop}
\end{figure}

RAI focuses on decreasing the dead node probability to increase expressive power, whereas ACI uses anti-correlated weights to improve the expressivity. As RAI and ACI increase the expressivity using different mechanisms, we explore the possibility of combining the two. We call it Random asymmetric anti-correlated initialization (RAAI). To prepare weights drawn from RAAI, we consider anti-correlated Gaussian weights and bias incoming to each node (like Equation \ref{eqn:corr_weights_dist}) and replace one randomly picked weight/bias with a random number drawn from a beta distribution. Note that the weights and biases reaching different nodes are uncorrelated. Like ACI, we observe three phases for RAAI. Numerical results presented in Figure \ref{fig:RAAI_forward_prop} suggest that the order to chaos boundary is around $\sigma^2_w = 0.9$, and the length diverges for $\sigma^2_w > 1.2$. Again, a qualitative picture of the phase diagram can be captured with a few additional assumptions over the mean-field approximation (see Appendix \ref{appendix:RAAI} for details).

In summary, RAAI has a chaotic phase like ACI, and a lower dead node probability like RAI, as can be checked numerically. As RAAI inherits the advantages of both strategies, we expect it to be a strong candidate for initializing ReLU networks. Table \ref{table:init} summarizes and compares different initialization schemes. In the following sections, we analyze the training dynamics and performance of RAAI and compare it with other initialization schemes.

\begin{table}[h]
  \caption{A comparison between different initialization schemes for ReLU networks. The dead node probabilities are calculated numerically for input signals drawn from the standard normal distribution.}
\label{table:init}
  \centering
  \begin{tabular}{llrcc}
    \toprule
    Initialization & $\sigma^2_w$ & $k$ & Chaotic & Dead node \\
	scheme         &        &  & phase & probability  \\
    \midrule
    He & 2.0 & 0.0 & No & 0.5\\[1pt] 
	ACI & 2.0 & 100.0 & Yes &  0.5\\[1pt]
	RAI & 0.36 & 0.0 & No & 0.36\\[1pt]
	RAAI & 0.92 & 100.0 & Yes &  0.36\\[1pt]
    \bottomrule%
  \end{tabular}
\end{table}

\section{Training tasks}

\label{sec:tasks}

This section describes various tasks used to analyze the dynamics and performance of different initialization schemes. We consider a variant of teacher-student setup  \cite{Seung1992}, in which a student ReLU network is trained with examples generated by an untrained teacher network. We consider three different tasks with varying complexities.
\begin{enumerate}
\item First, a standard teacher task, in which the training data is generated by a ReLU network of the same size as the student network, initialized with He initialization.

\item Next, we consider a simple teacher task in which the capacity of the teacher network is much lower than the student network. In many real data sets, the high-dimensional inputs lie in a low-dimensional manifold \cite{Goldt2020}, which motivates us to consider a simple teacher task.
We consider a single-layer ReLU network with $N = 10$ nodes, initialized with He initialization.

\item Lastly, we consider a complex teacher task, in which the complexity of the teacher network is more than the student network. We consider a teacher network with tanh activation of the same size as the student network initialized in the chaotic regime, $(\sigma^2_w, \sigma^2_b) = (1.5, 0)$ \cite{Poole2016}. A ReLU network initialized with symmetric distributions has half of the nodes dead. Therefore, it has a lower capacity than a tanh network of the same size. \end{enumerate}

We consider an $L = 10$ layered (in addition to input layer) student network with $N = 100$ nodes in each layer trained using SGD and Adam algorithms. We implement neural networks in Tensorflow (version $2.2.0$) \cite{tensorflow} and train them using $10^5$ training examples with a mean squared error loss, a mini-batch size of $1000$, and default parameters for the optimizers. The validation set contains $10^3$ examples.

\section{Comparison of learning dynamics for different initialization schemese}

\label{sec:train_RAAI}
This section compares the performance of RAAI with other initialization schemes listed in Table \ref{table:init} on tasks described in Section \ref{sec:tasks}.

\begin{figure}[h]
  \centering
  \includegraphics[width=0.9\linewidth]{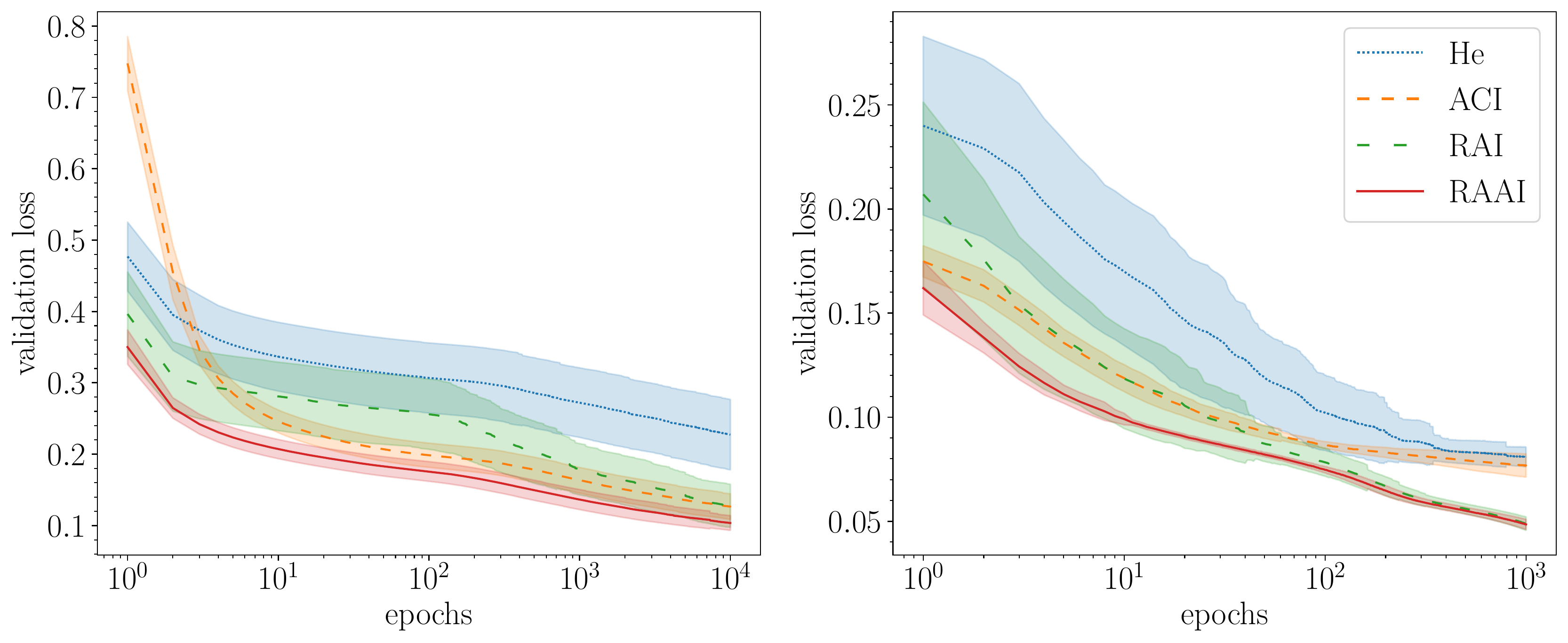}
    \caption{Average validation loss for ReLU networks trained on the standard teacher task with SGD (left) and Adam optimizer (right) for different initialization schemes. The shaded region shows the standard deviation around the average loss.}
    \label{fig:raai_train_standard}
\end{figure}

\paragraph{Standard teacher task} Figure \ref{fig:raai_train_standard} shows the average validation loss for the standard tasks trained with SGD and Adam algorithms. We observe that RAAI performs better than all other initialization schemes, whereas the relative performance of RAI and ACI depends on the optimization algorithm.

\begin{figure}[h]
  \centering
  \includegraphics[width=0.9\linewidth]{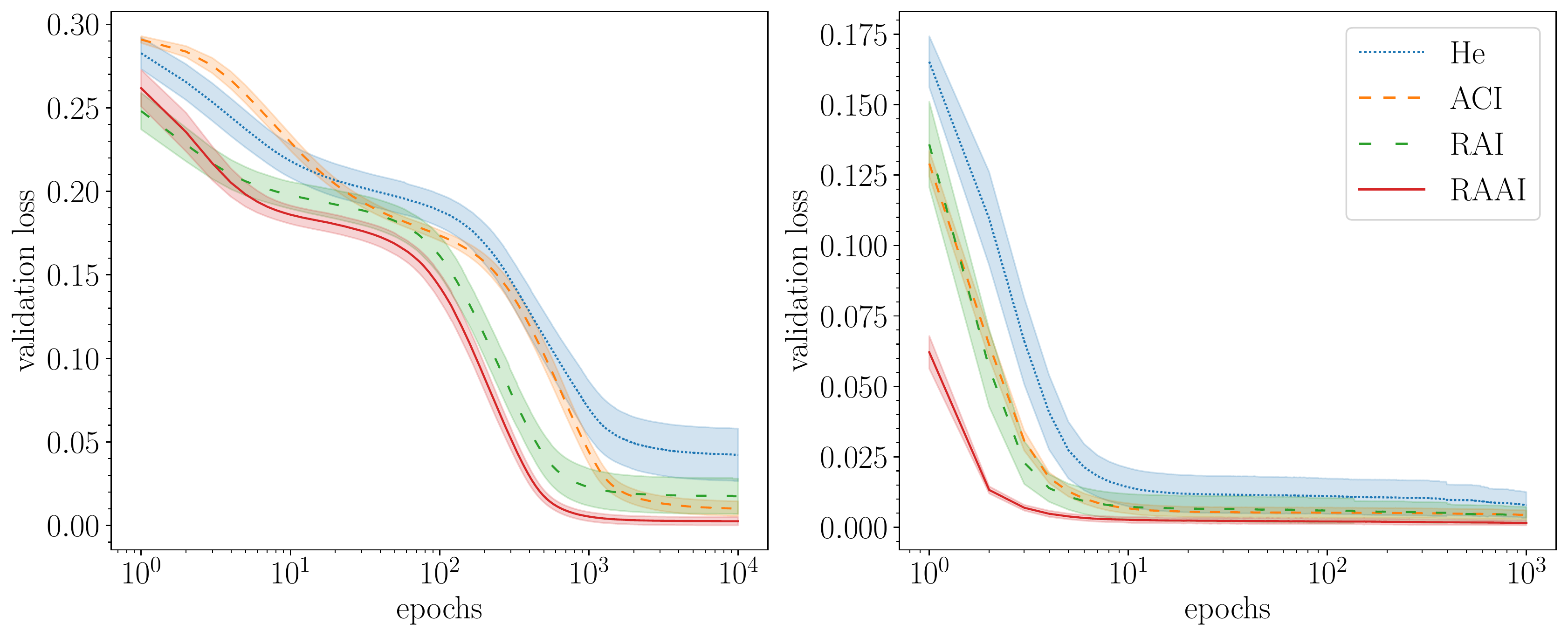}
    \caption{Average validation loss for ReLU networks trained on the simple teacher task with SGD (left) and Adam optimizer (right) for different initialization schemes. The shaded region shows the standard deviation around the average loss.}
    \label{fig:raai_train_easy}
\end{figure}

\paragraph{Simple teacher task} Figure \ref{fig:raai_train_easy} shows the average validation loss for the simple teacher task. Similar to the standard teacher task, RAAI performs better than or on par with other initialization schemes. In between RAI and ACI, the former performs better on using either optimizer.

\begin{figure}[h]
  \centering
  \includegraphics[width=0.9\linewidth]{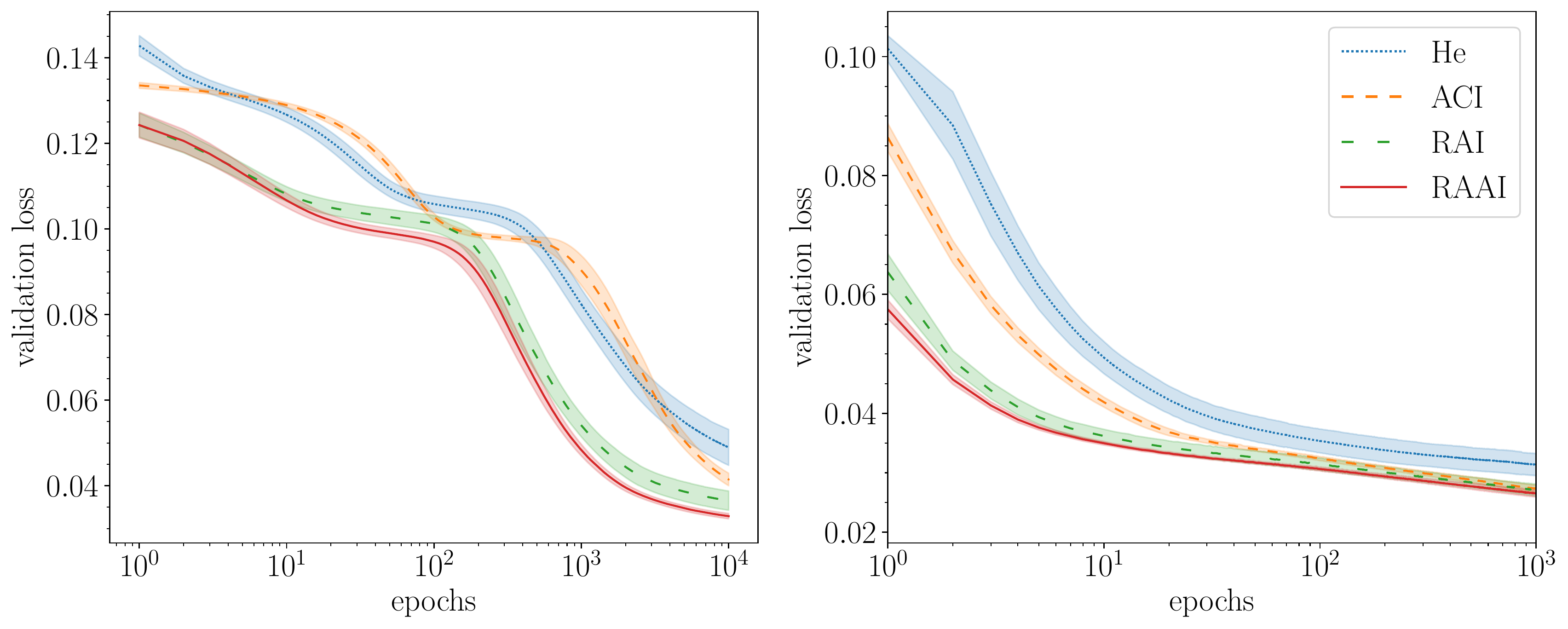}
    \caption{Average validation loss for ReLU networks trained on the complex teacher task with SGD (left) and Adam optimizer (right) for different initialization schemes. The shaded region shows the standard deviation around the average loss.}
    \label{fig:raai_train_hard}
\end{figure}

\paragraph{Complex teacher task} Figure \ref{fig:raai_train_hard} shows the average validation loss for the complex teacher task. We observe that for a complex teacher, ACI starts to perform worse when trained with SGD algorithm, whereas, RAAI faces no such problem and performs better or on par with all other initializations.

It may be of concern that initialization in an expressive subspace of weights might lead to overfitting. We trained a neural network with $L=20$ layers and found that ACI starts to overfit, but this can be avoided by reducing the value of the correlation strength $k$. This leads to another parameter to be tuned. On the other hand, RAAI does not show any overfitting signs and performs consistently better than other initialization schemes on training deeper networks. We believe that overfitting in ACI is related to correlations vanishing to zero at high variance (see Figure \ref{fig:AI_forward_prop}). In contrast, for RAAI, correlations do not vanish to zero at high variance (see Figure \ref{fig:RAAI_forward_prop}).

\section{Discussion and Conclusion}
\label{sec:discussion}
In this article, we analyzed the evolution of correlation between signals propagating through a ReLU network with correlated weights using the mean-field theory of signal propagation. 
Multiple studies show that ReLU networks with uncorrelated weights are biased towards computing simpler functions, but ReLU networks do perform complex tasks in practice.
Unlike ReLU networks with uncorrelated weights, ReLU networks with anti-correlated weights reaching a node have a chaotic phase where correlation saturates below unity. This suggests that such networks can exhibit higher expressivity. 
Although we have focused on the ReLU networks in this study, anti-correlation in weights may be useful in general.
Networks with other non-linear activation functions like tanh, SELU, and sigmoid have a chaotic phase even with uncorrelated weights. In these cases, the weight correlations may still help to tune the phase boundaries and expressivity of the networks.

We further investigated the possibility that ReLU networks with the enhanced expressivity may prove beneficial in faster learning. Comparison of training and test performance of networks in a range of teacher-student setups clearly showed that networks with anti-correlated weights learn faster.
While ACI shows better learning performance in general, it shows poor performance with SGD during an intermediate learning stage when the teacher network has a relatively higher capacity. We believe that this may be due to the system getting stuck in local minima. This is consistent with the absence of a similar regime on training with Adam optimizer.
On training deeper networks with ACI, we found that it overfits, but this can be avoided by fine-tuning correlation strength $k$.
We also investigated a possible improvement in training time from adding a regularization term in the loss function that favors anti-correlated weights, but our attempts did not show any systematic results.

We compared ACI with a recently proposed initialization scheme called RAI, which introduces a systematic asymmetry (around $0$) in the weights to decrease dead node probability. We find that the relative performance between RAI and ACI depends on the task and the optimization algorithm. RAI improves expressivity by reducing the dead node probability, whereas ACI achieves the same by inducing a chaotic phase. As RAI and ACI rely on different mechanisms, we explored a strategy of combining the two initialization schemes.
We analyzed the correlation properties of the combined scheme, which we call RAAI and found that it has a chaotic phase like ACI. 
We demonstrated that RAAI leads to faster training and learning than the best-known methods on various teacher tasks of a range of complexity.
For different initialization schemes, the behavior of the training dynamics at large epochs may depend on the optimizer and training data, however RAAI shows a definite advantage over other schemes when using the SGD optimizer, especially in early training epochs.
In addition to faster training, RAAI also shows no sign of overfitting and thus improves on the simpler strategy that relies only on anti-correlations.

\medskip
\section*{Acknowledgments} 
Dayal Singh would like to acknowledge the INSPIRE-SHE programme of Department of Science \& Technology, India. G J Sreejith acknowledges National Supercomputing Mission (Param Bramha, IISER Pune) for computational resources and financial support.
{
\small

\bibliographystyle{unsrt}
\bibliography{ref}
}

\appendix

\section{Derivation of the length and correlation maps for correlated weights}
\label{appendix:ACI}
This section derives the length and covariance maps for ReLU networks with correlated weights given by 
\begin{equation}
    \label{eqn:corr_weights_dist_appendix}
    P({\bm w}^l_1, {\bm w}^l_2, {\bm w}^l_3\dots) = \prod_i^{N_l} \frac{\exp\left(-\frac{1}{2} (\bm{w}^l_i)^T A^{-1} \bm{w}^l_i \right)} {\sqrt{(2\pi)^{N_{l-1}} |A| }},
\end{equation}
where the covariance matrix given by
\begin{equation*}
    \label{eqn:weight_covariance_appendix}
    A =  \frac{\sigma^2_w}{N_{l-1}} \left(\mathbb{I} - \frac{k}{1+k}\frac{J}{N_{l-1}} \right).
\end{equation*}
Here $\mathbb{I}$ is the identity matrix, $J$ is an all-ones matrix, and $k$ parameterizes the correlation strength. Positively correlated and anti-correlated regimes correspond to the regions $-1<k <0$ and $k > 0$, respectively, whereas, $k=0$ generates uncorrelated weights. For simplicity, we consider $N_l = N$ in all layers, but the results hold for all $N_l$, as long as it is large.
\subsection{Derivation of length map}

\label{appendix:length_map_AI}

To derive the length map, we follow the approach introduced by Ref. \cite{Poole2016}. Assuming self-averaging, we obtain the average value of the squared length of a signal, ${\bm s}^0=x$, after propagating to layer $l$ by considering an average over weights and biases between layer $l$ and $l-1$ 

\begin{equation}
  \label{eqn:A0}
      \centering
      \begin{split}
      q^l_h(x) &  = \frac{1}{N} \left< \sum_{i = 1}^N(h^l_{i}(x))^2 \right> = \frac{1}{N} \sum_{i=1}^N \sum_{j, m = 1}^{N} \left< w^l_{ij}w^l_{im} \right> \phi(h^{l-1}_{j}(x)) \phi(h^{l-1}_{m}(x)) + \left< (b^l_i)^2\right>  \\
      q^l_h(x) & = \frac{\sigma^2_w}{N} \sum_{j, m = 1}^{N} \left(\delta_{j, m} - \frac{k}{1+k} \frac{1}{N}\right) \phi(h^{l-1}_{j}(x)) \phi(h^{l-1}_{m}(x)) + \sigma^2_b,
      \end{split}
\end{equation}

where we have used $ \left< w^l_{ij}w^l_{im} \right> =  \frac{\sigma^2_w}{N} \left( \delta_{j, m} - \frac{k}{1+k} \frac{1}{N} \right)$ and $\left< (b^l_i)^2\right> = \sigma^2_b$.
For large $N$, each $h^{l-1}_i(x)$ is a weighted sum of a large number of correlated random variables, which converges to a zero-mean Gaussian with a variance $q^{l-1}_h(x)$. Replacing the average over $h^l_i$ at layer $l-1$ by a Gaussian distribution to get the general form of the recursive map. This average corresponds to averaging over all the weights and biases upto layer $l-1$.

\begin{equation*}
    \label{eqn:A1}
    q^l_h(x) = \sigma^2_w \int Dz\;\phi\left(\sqrt{q^{l-1}_h(x)} \; z \right)^2  -\sigma^2_w \frac{ k}{1+k} \left[ \int Dz \; \phi\left(\sqrt{q^{l-1}_h(x)} \;z \right)  \right]^2 + \sigma^2_b,
\end{equation*}

where $Dz$ is the standard normal distribution.
For the second term in the Equation \ref{eqn:A0}, we have used the fact that for different nodes $m \neq j$, $h^l_j(x)$ and $h^l_m(x)$ are uncorrelated random variables and ignored $\mathcal{O}(1/N)$ terms. Note that the weights and biases reaching two different nodes are uncorrelated. 
For a ReLU activation, we can perform the integrals to get the exact form of the recursive relation between $q^l_h(x)$ and $q^{l-1}_h(x)$

\begin{align}
    \label{eqn:length_map_AI}
    &q^l_h(x) = \frac{\sigma^2_w }{2}  \left( 1 - \frac{k}{1+k} \frac{1}{\pi} \right)q^{l-1}_h(x)  + \sigma^2_b.
\end{align}.

\subsection{Derivation of the covariance map}
\label{appendix:cov_map_AI}
The covariance map can be derived similarly by considering an average over the weights and biases
\begin{align*}
     & q^l_h(x_1, x_2)  = \frac{1}{N} \left< \sum_{i = 1}^N  h^l_{i}(x_1) h^l_{i}(x_2) \right> = \sum_{j, m = 1}^{N} \left< w^l_{ij}w^l_{im} \right> \phi(h^{l-1}_{j}(x_1)) \phi(h^{l-1}_{m}(x_2)) + \left< (b^l_i)^2\right>  \\
      &q^l_{h}(x_1, x_2) = \frac{\sigma^2_w}{N} \sum_{j, m = 1}^{N} \left(\delta_{j, m} - \frac{k}{1+k} \frac{1}{N}\right) \phi(h^{l-1}_{j}(x_1)) \phi(h^{l-1}_{m}(x_2)) + \sigma^2_b,
\end{align*}

and then replacing the sum over neurons in the previous layer with an integral with a Gaussian measure.
For large $N$, the joint distribution of $h^l_{j}(x_1) $ and $h^l_{m}(x_2)$ will converge to a two-dimensional Gaussian distribution with a covariance matrix

\begin{equation*}
\Sigma_{l-1} = \begin{bmatrix}q^{l-1}_h(x_1)  &  q^{l-1}_h(x_1, x_2) \\ q^{l-1}_h(x_1, x_2) & q^{l-1}_h(x_2)  
\end{bmatrix}.
\end{equation*}

The correlations among $h^l_{j}(x_1) $ and $h^l_{m}(x_2)$ are induced as the two signals are propagating through the same network. Propagating this joint distribution across one layer, we obtain the iterative map
\begin{equation*}
    \label{eqn:A3}
    \begin{split}
         & q^l_{h}(x_1, x_2) =   \sigma^2_w \int Dz_1 Dz_2 \phi(u_1)\phi(u_{12})  - \sigma^2_w \frac{k}{k+1} \int Dz_1 Dz_2 \;  \phi(u_1) \phi(u_2) + \sigma^2_b \\
    & u_1 = \sqrt{q^{l-1}_h(x_1)}z_1, \; u_{12} = \sqrt{q^{l-1}_h(x_2)} \left[ c^{l-1}_h \;z_1 + \sqrt{1 - (c^{l-1}_h)^2} \; z_2 \right]u_2 = \sqrt{q^{l-1}_h(x_2)}z_2, 
    \end{split}
\end{equation*}
    
where $c^l_h = \nicefrac{q^l_h(x_1, x_2)}{\sqrt{q^l_h(x_1)q^l_h(x_2)}}$ is the correlation coefficient, and $Dz_1, Dz_2$ are standard normal Gaussian distributions. Again, in the second part of the above equation, we have used the fact that for $j \neq m$, $h^l_{j}(x_1)$ and $h^l_{m}(x_2)$ are uncorrelated random variables and have ignored $\mathcal{O}(1/N)$ terms. Further, we can perform the integrals for ReLU networks to get the exact recursive map

\begin{align}
&q^l_{h}(x_1, x_2) =  \frac{\sigma^2_w }{2} \left( f(c^{l-1}_h) - \frac{ k}{1+k} \frac{1}{\pi} \right)\sqrt{q^{l-1}_h(x_1)q^{l-1}_h(x_2)} + \sigma^2_b\\
     & f(c^{l-1}_h) =  \frac{c^{l-1}_h}{2} + \frac{c^{l-1}_h}{\pi}  \text{sin}^{-1}(c^{l-1}_h)  + \frac{1}{\pi}\sqrt{1 - (c^{l-1}_h)^2}.
\end{align}

\section{Training with anti-correlated vs. positively correlated initialization}

\label{sec:train_ACI}

This section compares the training dynamics and performance of ReLU networks initialized with different weight correlation strengths on tasks described in Section 4. We utilize the increased expressivity in ReLU networks with anti-correlated weights (ACI) at the initial training phase and compare its training dynamics with He initialization and positively correlated weight initialization. We observe that ACI provides a definite advantage over the other two initialization schemes. We also find that positively correlated weight ReLU networks train slower than He initialization, suggesting that anti-correlation may develop in weights during training. We choose three different correlation strengths; $k = 100$ induces anti-correlated weights, $k = -0.5$ produces positively correlated weights, and lastly, $k = 0$ corresponds to uncorrelated weights (He initialization). We train networks with two different optimization algorithms, SGD and Adam. For SGD, we train for $10^4$ epochs, and for Adam, we train for $10^3$ epochs. 

\begin{figure}[h]
  \centering
  \includegraphics[width=1\linewidth]{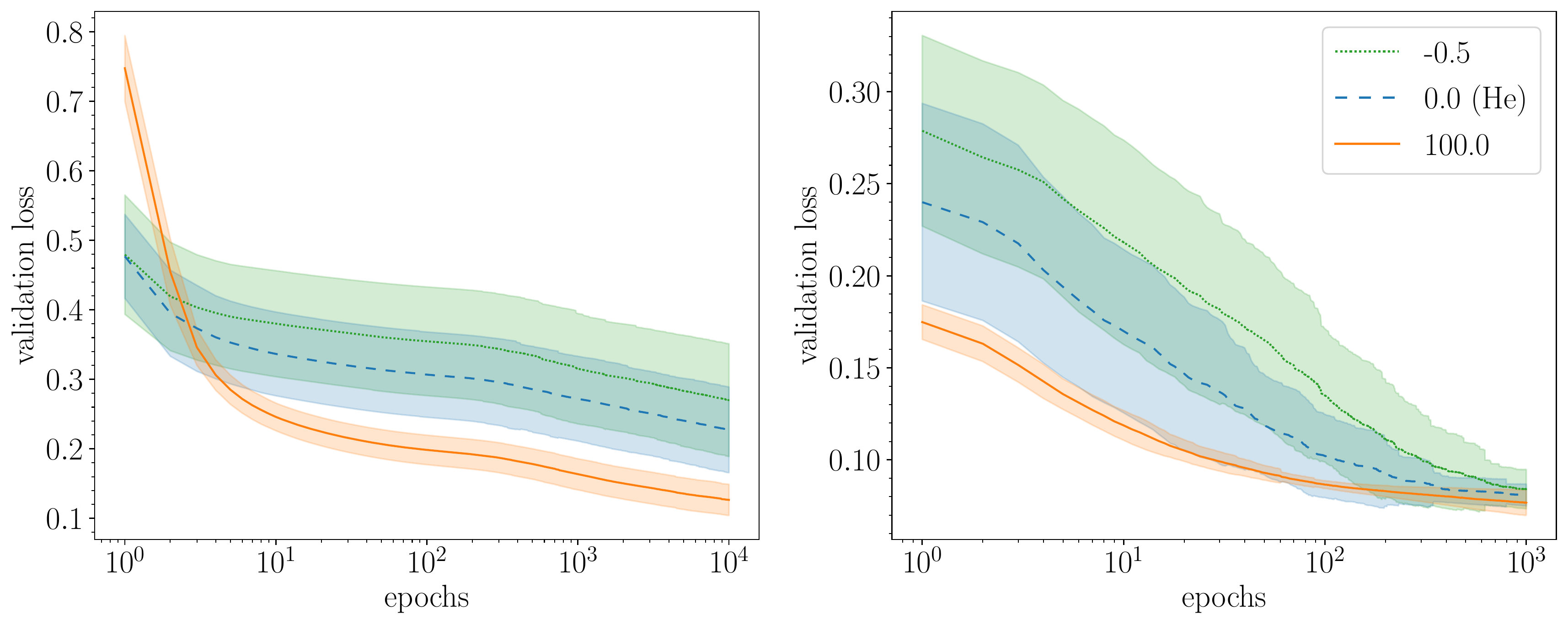}
    \caption{Average validation loss for ReLU networks trained on the standard teacher task with SGD (left) and Adam optimizer (right) for different weight correlations strengths.}
    \label{fig:train_standard}
\end{figure}

\paragraph{Standard teacher task} Figure \ref{fig:train_standard} shows the average validation loss for different correlation strengths trained using SGD and Adam algorithms. We observe that ReLU networks initialized with ACI train faster than He initialization. In contrast, ReLU networks with positively correlated weights train slower than He initialization.

\begin{figure}[h]
  \centering
  \includegraphics[width=1\linewidth]{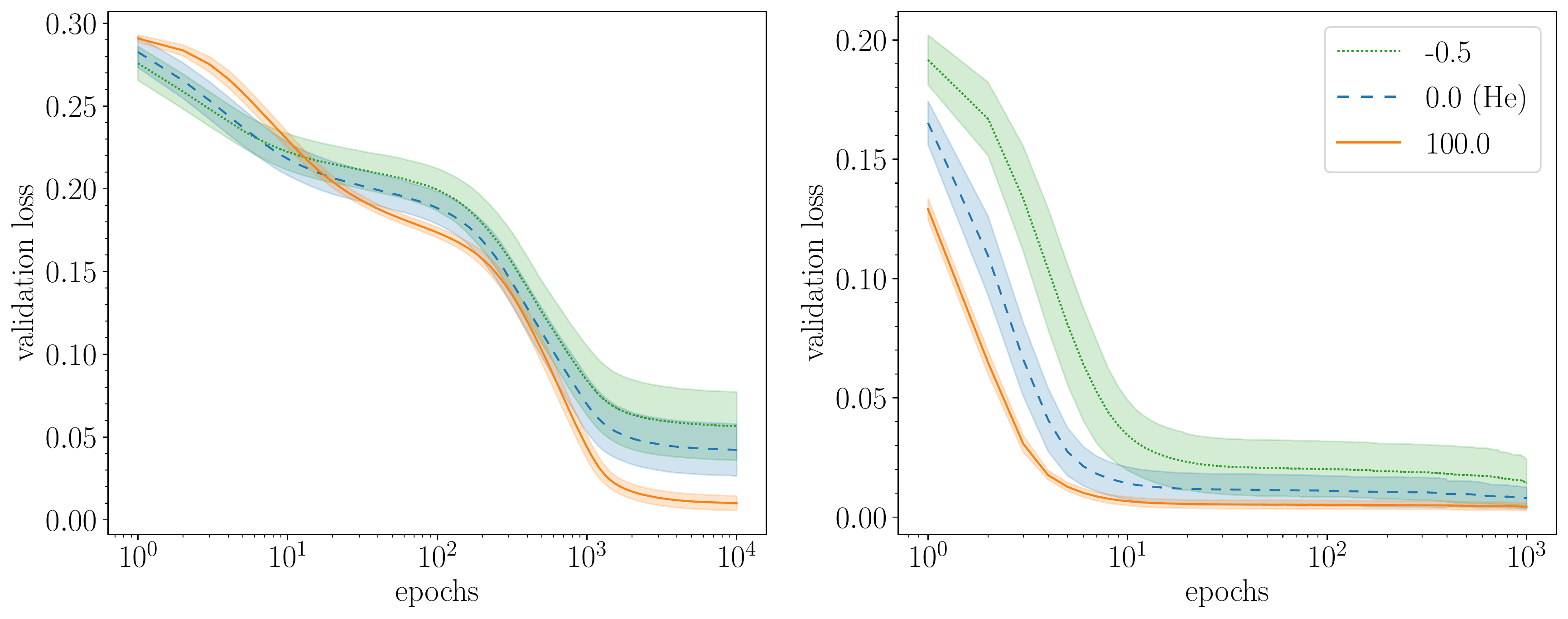}
    \caption{Average validation loss for ReLU networks trained on the simple teacher task with SGD (left) and Adam (right) optimizer for different weight correlations strengths.}
    \label{fig:train_easy}
\end{figure}

\paragraph{Simple teacher task} Figure \ref{fig:train_easy} shows the average validation loss for the simple teacher task. We observe similar qualitative results as in the standard teacher task. For SGD, we observe an initial linear region in which all initialization schemes perform equally; however, at large epochs, ACI shows a definite advantage over other initializations.

\begin{figure}[h]
  \centering
  \includegraphics[width=1\linewidth]{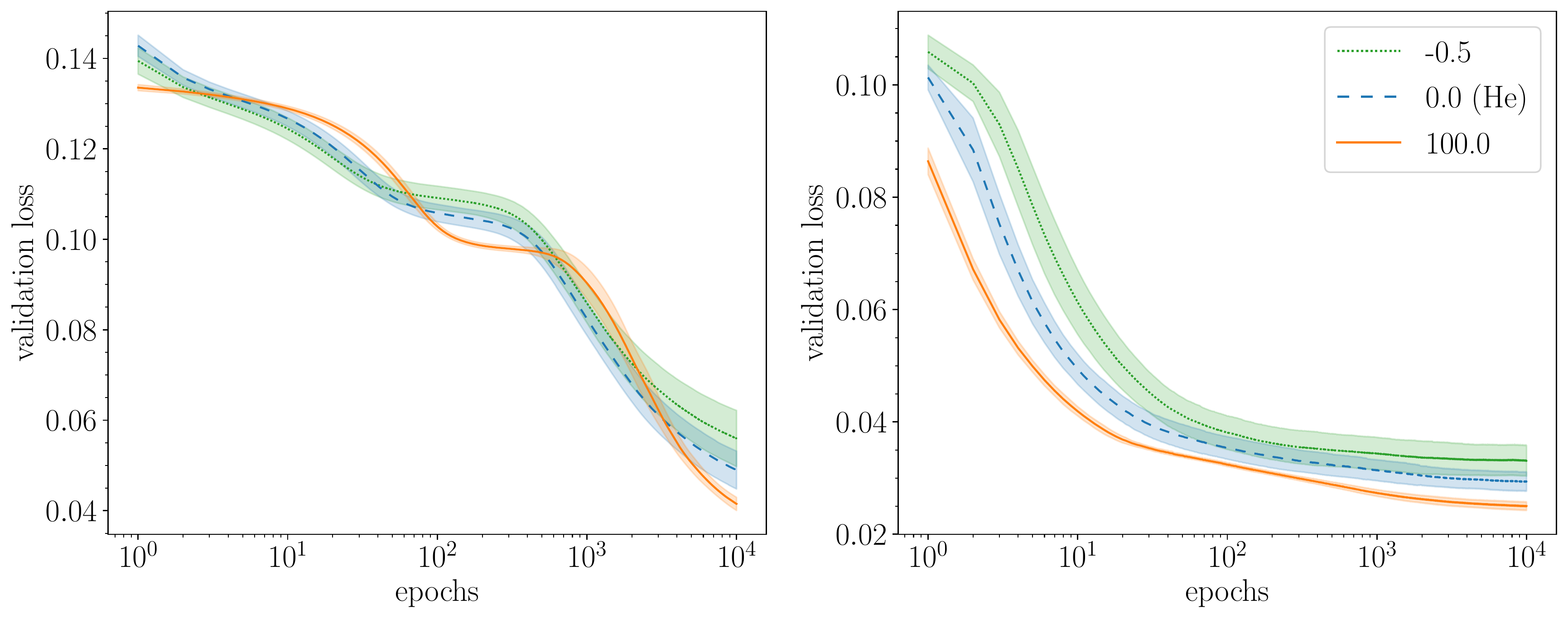}
    \caption{Average validation loss for ReLU networks trained on the complex teacher task with SGD (left) and Adam (right) optimizer for different weight correlations strengths.}
    \label{fig:train_hard}
\end{figure}

\paragraph{Complex teacher task} Figure \ref{fig:train_hard} shows the average validation loss for a complex teacher task. For some intermediate regions, ACI performs worse than other initializations on training with SGD. The regions where ACI performs poorly shift depending on the complexity of the task.

\section{Derivation of the length and covariance map for RAI}

 To draw weights from the RAI distribution, we first initialize the weights and bias incoming to each node with a Gaussian distribution $\mathcal{N}(0, \nicefrac{\sigma^2_w}{N})$. Next, we replace one weight or the bias incoming to each neuron by a random variable from beta distribution (see Ref. \cite{Lu2020} for details). The weights and bias are treated on an equal footing. Thus, to simplify the notations, we incorporate the bias in the weight matrix by introducing a fictitious additional node with a constant value of one, i.e., 

\begin{equation*}
    \mathbf{s}^l(x) = [\phi(\mathbf{h}^l(x)), 1].
\end{equation*}

 The evolution equation is now given by

\begin{equation*}
    \label{eqn:evolution_RAI}
    \mathbf{h}^l(x) = \mathbf{W}^l \cdot \mathbf{s^{l-1}}(x) .
\end{equation*}

 It is easier to track the evolution using the activation instead of the pre-activations. So we define a few covariance matrices which will come in handy

\begin{align*}
    & q^l_{s}(x_1, x_2) = \frac{1}{N+1} \sum_{i = 0}^{N} s^l_{i}(x_1) s^l_{i}(x_2)  \\
    & q^l_{-k_j^l}(x_1,  x_2) = \frac{1}{N} \sum^{N}_{t \neq k_j^l} s^l_{t}(x_1)s^l_{t}(x_2) \; ,
\end{align*}

where $k^l_j$ tags variables associated with the special weight. We will use the notations $q^l_s(x) = q^l_s(x, x)$ and $q^l_{-k}(x) = q^l_{-k}(x, x)$. The corresponding correlation coefficients are given by,

\begin{align*}
    & c^l_{s} = \frac{q^l_{s}(x_1, x_2)}{\sqrt{q^l_{s}(x_1) \; q^l_{s}(x_2)}} \\
    & c^l_{-k_j^l} = \frac{q^l_{-k_j^l}(x_1, x_2)}{\sqrt{q^l_{-k_j^l}(x_1) \; q^l_{-k_t^l}(x_2)}}
\end{align*}

\subsection{Derivation of the length map for RAI}
\label{appendix:length_map_RAI}

Given $h^{l-1}(x)$ and considering weights between layers $l$ and $l-1$, we can view $h^l_j(x)$ as a random variable

\begin{equation*}
    h^l_j(x) =   \sigma_w \; \sqrt{q^{l-1}_{-k^{l-1}_j}(x)}\; z + s^{l-1}_{k_j^{l-1}}(x) \;u ,
\end{equation*}

where $z \sim \mathcal{N}(0, 1)$ and $u \sim \beta(2, 1)$. By applying the activation function and squaring it, we obtain
\begin{equation*}
    \phi(h^l_{j}(x))^2 =  \phi\left(\sigma_w \; \sqrt{ q^{l-1}_{-k^{l-1}_j}(x)}\; z + s^{l-1}_{k_j^{l-1}}(x) \; u \right)^2.
\end{equation*}

Next, we take an average over the weights and the special weight ( average denoted by $<.>$) to get

\begin{equation*}
    \left< \phi(h^l_{j}(x))^2 | h^{l-1}(x) \right> = \sum_{k^{l-1}_j = 0}^N \frac{1}{N+1} \int dz \;du \; f(z)\; g(u) \; \phi\left(\sigma_w \sqrt{q^{l-1}_{-k^{l-1}_j}(x)} \; z + s^{l-1}_{k_j^{l-1}}(x) \; u \right)^2, 
    \end{equation*}
    
where $g(u) \sim \beta(2, 1)$ distribution, and $f(z) \sim \mathcal{N}(0, 1)$. We can take a sum over all nodes and re-write the equation in terms of the overlap

\begin{equation}
    \label{eqn:length_map_RAI}
    \left< q^l_s(x) | h^{l-1}(x) \right> = \frac{1}{N + 1}\left[ 1 + \sum_{j = 1}^N \sum_{k^{l-1}_j = 0}^N \frac{1}{N+1} \int dz du \; f(z) g(u) \; \phi\left(\sigma_w \sqrt{q^{l-1}_{-k^{l-1}_j}(x)} \; z + s^{l-1}_{k_j^{l-1}}(x) \; u \right)^2 \right].
    \end{equation}

\subsection{Derviation of the covariance map for RAI}
\label{appendix:RAI_cov_map}
The covariance map can be derived similarly, with a key difference of covariance between the pre-activations. For two input signals ${\bm s}^0 = x_1$ and ${\bm s}^0=x_2$, the covariance map reads

\begin{equation}
    \label{eqn:cov_map_RAI}
    \begin{split}
    & \left< q^l_s(x_1, x_2) |  h^{l-1}(x_1), h^{l-1}(x_2) \right> = \frac{1}{N+1} \left[  1 + \sum_{j = 1}^N \sum_{k^{l-1}_j = 0}^N \frac{1}{N+1} \int dy_1 dy_2 du \; f(y_1, y_2)\; g(u)  \right. \times \\
    & \times  \;  \left.\phi\left(\sigma_w \sqrt{q^{l-1}_{-k^{l-1}_j}(x_1)} \; y_1 + s^{l-1}_{k_j^{l-1}}(x_1) \; u \right) \phi\left(\sigma_w \sqrt{q^{l-1}_{-k^{l-1}_j}(x_2)} \; y_2 + s^{l-1}_{k_j^{l-1}}(x_2) \; u \right) \right],
    \end{split}
    \end{equation}

where $f(y_1, y_2)$ is the joint Gaussian distribution of $y_1$ and $y_2$, with a covariance matrix given by

\begin{equation*}
    \label{eqn:cov_matrix_RAI}
    \Sigma^{l-1}_{k^l_j} = 
    \begin{bmatrix}
     q^{l-1}_{-k^{l-1}_j}(x_1) & q^{l-1}_{-k^{l-1}_j}(x_1, x_2)\\
    q^{l-1}_{-k^{l-1}_j}(x_1, x_2) & q^{l-1}_{-k^{l-1}_j}(x_2)
     \end{bmatrix}.
\end{equation*}

We can re-write Eqn. \ref{eqn:cov_map_RAI} in terms of $ c^l_{-k_j^l}$

\begin{equation*}
    \begin{split}
      \left< q^l_s(x_1, x_2)  |  h^{l-1}(x_1), h^{l-1}(x_2) \right> =   \frac{1}{N + 1}& \left[ 1 + \sum_j \sum_{k^{l-1}_j} \frac{1}{N+1} \int dz_1 \;dz_2 \;du \; f(z_1) f(z_2)\; g(u)  \right. \times \\ 
    & \times  \; \phi\left(\sigma_w \sqrt{q^{l-1}_{-k^{l-1}_j}(x_1)} \; z_1 + s^{l-1}_{k_j^{l-1}}(x_1) \; u \right) \times \\
    & \times \left. \phi\left(\sigma_w \sqrt{q^{l-1}_{-k^{l-1}_j}(x_2)} \left[  c^l_{-k_j^l} z_1 + \sqrt{1 - ( c^l_{-k_j^l})^2} \;z_2 \right] + s^{l-1}_{k_j^{l-1}}(x_2) \; u \right) \right],
    \end{split}
    \end{equation*}

where $f(z_1) \sim f(z_2) \sim \mathcal{N}(0, 1)$ are standard Gaussian distributions. As suggested by Ref. \cite{Poole2016}, we can find the fixed point of the correlation map under the assumption that the length $q^l_h(x)$ has reached its fixed point. It can be checked that $c^l_h = 1$ is a fixed point of the correlation map.
\section{Stability of the fixed points for the length and correlation maps for RAI}

\subsection{Stability of the fixed point for the length map for RAI}

The derivation of the analytical form of the length map (Eqn. \ref{eqn:length_map_RAI}) is difficult, and only bounds to the map have been derived (see Ref. \cite{Lu2020}). Inspired by the analytical form of the length map for the anti-correlated initialization and the analysis done by Ref. \cite{Lu2020}, we assume that the length map has a linear dependence on $q^{l-1}_s(x)$. Under this assumption, we can find the stability of the fixed point of the length map by taking a derivative with respect to $q^{l-1}_s(x)$. Yet another problem exists. While taking a derivative, we have to encounter derivatives of the form 
\begin{equation*}
\frac{\partial s^{l-1}_{-k^{l-1}_j}(x)}{\partial q^{l-1}_h(x)}.
\end{equation*}
To simplify the calculations further, we employ a mean-field type approach by approximating $q^{l-1}_{-k^l_j}(x)$ and $s^{l-1}_{k^{l-1}_j}$ by $q^{l-1}_s(x)$. Note that we can also approximate $s^{l-1}_{k^l_j}$ by its mean value, giving the same qualitative results.
This simplifies Eqn. \ref{eqn:length_map_RAI} to

\begin{equation*}
    \left< q^l_s(x) | h^{l-1}(x) \right> = \frac{1}{N + 1}\left[ 1 + (N+1) \int dz \;du \; f(z)\; g(u) \; \phi\left(\sqrt{q^{l-1}_{s}(x)} (\sigma_w   z +  u) \right)^2 \right].
    \end{equation*}

To find the fixed point of the length map, we take a derivative wrt $q^{l-1}_s(x)$ to get the condition for stability of the fixed point $q^*$. We denote this derivative by $\zeta_{q^*}$. It separates the dynamics into two phases \textemdash a bounded phase when $\zeta_{q^*} < 1$, and an unbounded phase when $\zeta_{q^*} > 1$.

\begin{align*}
     \zeta_{q^*} = & \frac{\partial q^l_s(x)}{\partial q^{l-1}_s(x)} \Bigr\rvert_{q^{l-1}_s(x)= q^*}  \\
     \zeta_{q^*} = & \frac{\partial}{\partial q^{l-1}_s(x)} \int dz \;du \; f(z)\; g(u) \; \phi\left( \sqrt{q^{l-1}_s(x)} (\sigma_w   z +  u) \right)^2 \\
     \zeta_{q^*} = & \frac{1}{\sqrt{q^{l-1}_s(x)}} \int dz \;du \; f(z)\; g(u) \;(\sigma_w z + u) \phi'\left( \sqrt{q^{l-1}_s(x)} (\sigma_w   z +  u) \right) \phi\left( \sqrt{q^{l-1}_s(x)} (\sigma_w   z +  u) \right) 
    \end{align*}
\begin{equation}
    \label{eqn:zeta_RAI}
     \zeta_{q^*} =  \sigma^2_w \int dz \;du \; f(z)\; g(u) \; \left[\phi'\left(  \sigma_w   z +  u  \right)\right]^2 + \sigma_w \int dz \;du \; f(z)\; g(u)\phi'\left( \sigma_w   z +  u  \right)\phi\left( \sigma_w   z +  u  \right)
\end{equation}
where we have used the fact that for $a >0$, $\phi(ax) = a \;  \phi(x)$. On evaluating the integral, we find that $\zeta_{q^*} = 1$ when $\sigma^2_w = 0.56$. This critical value underestimates the numerical value obtained in Figure \ref{fig:RAI_forward_prop}.

\subsection{Stability of the fixed point for the correlation map for RAI}
\label{appendix:stability_corr_map_RAI}
Under the assumption, $q^l_s(x) \to q^*$, the correlation map has a fixed point $c^*_{s} = 1$, and its stability is given by $\chi_1 = \nicefrac{\partial c^l_s}{ \partial c^{l-1}_s}$ evaluated at $c^{l-1}_s = 1$. But again, we get into the difficulties mentioned in the previous section, and we employ the same assumptions to arrive at a tractable equation for the correlation map

\begin{equation*}
    \begin{split}
      & \left< c^l_s  |  h^{l-1}(x_1), h^{l-1}(x_2) \right> =   \frac{1}{q^*_s(x)} \frac{1}{N + 1} \left[ 1 + N \int dz_1 \;dz_2 \;du \; f(z_1) f(z_2)\; g(u)  \right. \times \\ 
    & \times  \; \left. \phi\left( \sqrt{q^*_s(x)}  (\sigma_w\; z_1 + u) \right) \phi\left( \sqrt{q^{*}_{s}(x)} \left[  c^{l-1}_s \; \sigma_w \; z_1 + \sqrt{1 - ( c^{l-1}_s)^2}\; \sigma_w \;z_2 + u \right]  \right) \right],
    \end{split}
    \end{equation*}

Next, we take a derivative to get the condition for the stability of the fixed point $c^*_h= 1$

\begin{align*}
     \chi_1 = & \frac{\partial c^l_h}{\partial c^{l-1}_h} \Bigr\rvert_{c^{l-1}_h= 1}  \\
     \chi_1 = &  \frac{1}{q^*_h(x)} \frac{\partial}{ \partial c^{l-1}_h} \int dz_1 dz_2 du f(z_1) f(z_2) g(u) \phi\left( \sqrt{q^*_s(x)}  (\sigma_w\; z_1 + u) \right) \times \\
    & \times   \phi\left( \sqrt{q^{*}_{s}(x)} \left[  c^{l-1}_s \; \sigma_w \; z_1 + \sqrt{1 - ( c^{l-1}_s)^2}\; \sigma_w \;z_2 + u \right]  \right) \Bigr\rvert_{c^{l-1}_h= 1} 
    \end{align*}

\begin{equation}
    \label{eqn:chi1_rai}
    \chi_1 =  \sigma^2_w \int dz \; du \; f(z) \; g(u) \; \left[ \phi'(\sigma_w z + u)\right]^2.
\end{equation}
The above equation is the same as the first term we obtained in the condition for the stability of the length map (Eqn. \ref{eqn:zeta_RAI}). We obtain a critical value of $\sigma^2_w = 1.41$ by solving for $\chi_1 = 1$. 
We observe that the critical point for the length is smaller than the critical point of the correlation coefficient, and from our experience with ReLU networks calculations, we expect RAI to have an ordered phase only, which is confirmed by numerical results shown in Figure \ref{fig:RAI_forward_prop}.

\section{Derivation of length and correlation map for RAAI and stability conditions}
\label{appendix:RAAI}
\subsection{Derivation for the length map for RAAI and the stability condition}

Similar to the previous section, we can view $h^l_j(x)$ as a random variable
\begin{equation*}
    h^l_j(x) = \sigma_w  \sqrt{\Tilde{q}^{l-1}(x)} \;z + s^{l-1}_{-k^{l-1}_j}(x) \; u ,
\end{equation*}

where $\Tilde{q}^{l-1}(x) = q^{l-1}_{-k^{l-1}_j}(x) \left(1 - \frac{k}{1+k} \frac{1}{\pi} \right)$. Then, we can re-define $\sigma_w$ as

\begin{equation*}
    \Tilde{\sigma}^2 = \sigma^2_w \left( 1 - \frac{k}{1+k}\frac{1}{\pi} \right),
\end{equation*}
which yields,

\begin{equation*}
    h^l_j(x) =  \Tilde{\sigma}  \sqrt{q^{l-1}_{-k^{l-1}_j}(x)} \;z + s^{l-1}_{-k^{l-1}_j}(x) \; u ,
\end{equation*}

Now, the entire analysis goes through as Appendix \ref{appendix:length_map_RAI}, just with a re-definition of the variance. Now, we can read off the stability condition for the fixed point of the length map

\begin{equation}
    \label{eqn:zeta_RAAI}
     \zeta_{q^*} =  \Tilde{\sigma}^2_w \int dz \;du \; f(z)\; g(u) \; \left[\phi'\left( \Tilde{\sigma}   z +  u  \right)\right]^2 + \Tilde{\sigma} \int dz \;du \; f(z)\; g(u)\phi'\left( \Tilde{\sigma} z +  u  \right)\phi\left( \Tilde{\sigma}  z +  u  \right)
\end{equation}

On solving the equations numerically, we observe that the length is bounded when $\sigma^2_w < 1.75$, which overestimates the numerical value observed in Figure \ref{fig:RAAI_forward_prop}.

\subsection{Derivation for the correlation map for RAAI and the stability condition}

The correlation map for RAAI can be derived similar to RAI (Appendix \ref{appendix:RAI_cov_map}), with a key difference being the covariance matrix. The covariance matrix, in this case, is

\begin{equation*}
    \label{eqn:cov_matrix_RAAI}
    \Sigma^{l-1}_{k^l_j}(x_1, x_2) = 
    \begin{bmatrix}
     q^{l-1}_{-k^{l-1}_j}(x_1)- \frac{k}{1+k} \left(m^l_{-k^{l-1}_j}(x_1)\right)^2 & q^{l-1}_{-k^{l-1}_j}(x_1, x_2) - \frac{k}{1+k} m^l_{-k^{l-1}_j}(x_1)m^l_{-k^{l-1}_j}(x_2)\\
    q^{l-1}_{-k^{l-1}_j}(x_1, x_2) - \frac{k}{1+k} m^l_{-k^{l-1}_j}(x_1)m^l_{-k^{l-1}_j}(x_2) & q^{l-1}_{-k^{l-1}_j}(x_2) - \frac{k}{1+k} \left(m^l_{-k^{l-1}_j}(x_2)\right)^2
     \end{bmatrix}.
\end{equation*}

Again, we can check that $c^*_s=1$ is a fixed point of the dynamics. The stability of the fixed point under the assumptions considered in Appendix \ref{appendix:stability_corr_map_RAI} determines the order to chaos boundary at $\sigma^2_w = 1.41$. The critical value overestimates the numerical results presented in Figure \ref{fig:RAAI_forward_prop}.

Note that instead of approximating $s^{l-1}_{k^{l-1}_j}$ by its RMS value $q^{l-1}_s(x)$, we can also approximate it by its mean value. In this case, we observe similar qualitative results. In Table \ref{table:approx}, we compare the boundaries predicted by the RMS and mean approximations.

\begin{table}[h]
\caption{A comparison between the decision boundaries obtained by approximating $s^{l-1}_{k^{l-1}_j}$ by its RMS and mean value. The RMS approximation underestimates the length boundary for RAI, whereas it overestimates both the phase boundaries for RAAI. On the other hand, the mean approximation overestimates the phase boundaries for RAI and RAAI both.}
\label{table:approx}
\centering
\begin{tabular}{lccc}
\toprule
Approximation & $(\sigma^2_w)_q (RAI)$ & $(\sigma^2_w)_c(RAAI)$ & $(\sigma^2_w)_q(RAAI)$ \\
\midrule
RMS & 0.57 & 1.41 & 1.75\\[1pt] 
Mean & 0.85 & 1.46 & 1.89\\[1pt]
\bottomrule
\end{tabular}
\end{table}

\section{Code to generate weights drawn from RAAI distribution}
\label{appendix:code}
\begin{lstlisting}[language=Python]
import numpy as np

def RAAI(fan_in, fan_out, k = 100, variance_weights = 0.9):
    """Randomized Asymmetric Anti-correlated Initializer (RAAI)
    Arguments:  
    fan_in -- the number of neurons in the previous layer
    fan_out -- the number of neurons in the next layer
    corr -- correlation strength for the Gaussian weights 
    variance_weights -- variance of the weights
    Returns:    
    W, b -- weight and bias matrices with shape(fan_in, fan_out), and (fan_out, )
    """
    corr = k/(1+k)
    mean = np.zeros(fan_in + 1)
    J = np.ones((fan_in + 1, fan_in + 1))
    cov = (np.identity(fan_in + 1) - J*(corr/(fan_in +1)) )*variance_weights/fan_in
    P = np.random.multivariate_normal(mean = mean, cov = cov, size = (fan_out)) 
    for j in range(P.shape[0]):
        k = np.random.randint(0, high = fan_in + 1)
        P[j, k] = np.random.beta(2, 1)
    W = P[:, :-1].T
    b = P[:, -1]
    return  W.astype(np.float32), b.astype(np.float32)

\end{lstlisting}

\end{document}